\documentclass{article}
\usepackage{ijcai22}

\usepackage{times}
\usepackage{soul}
\usepackage{url}
\usepackage[hidelinks]{hyperref}
\usepackage[utf8]{inputenc}
\usepackage[small]{caption}
\usepackage{graphicx}
\usepackage{amsmath}
\usepackage{amsthm}
\usepackage{booktabs}
\usepackage{algorithm}
\usepackage{algorithmic}
\usepackage{setspace}
\usepackage{multirow}
\usepackage{multicol}
\usepackage{makecell}
\usepackage{booktabs}
\usepackage{xcolor}
\usepackage{tablefootnote}
\title{Tackling Math Word Problems with Fine-to-Coarse Abstracting and Reasoning}

\author{
Ailisi Li$^1$
\and
Xueyao Jiang$^1$\and
Bang Liu$^{2}$\and
Jiaqing Liang$^{1}$\And
Yanghua Xiao$^1$
\affiliations
$^1$Fudan University\\
$^2$Mila \& DIRO, Universit{\'e} de Montr{\'e}al, Canada CIFAR AI Chair
\emails
\{alsli19, xueyaojiang19\}@fudan.edu.cn.com,
bang.liu@umontreal.ca,
l.j.q.light@gmail.com,
shawyh@fudan.edu.cn
}

\begin{document}

\maketitle

% \doublespacing

\begin{abstract}
Math Word Problems (MWP) is an important task that requires the ability of understanding and reasoning over mathematical text.
Existing approaches mostly formalize it as a generation task by adopting Seq2Seq or Seq2Tree models to encode an input math problem in natural language as a global representation and generate the output mathematical expression.
Such approaches only learn shallow heuristics and fail to capture fine-grained variations in inputs. 
In this paper, we propose to model a math word problem in a fine-to-coarse manner to capture both the local fine-grained information and the global logical structure of it. Instead of generating a complete equation sequence or expression tree from the global features, we iteratively combine low-level operands to predict a higher-level operator, abstracting the problem and reasoning about the solving operators from bottom to up. 
Our model is naturally more sensitive to local variations and can better generalize to unseen problem types. 
Extensive evaluations on Math23k and SVAMP datasets demonstrate the accuracy and robustness of our method.\footnote{Code and data used in this paper will be released after publication.}
\end{abstract}

\section{Introduction}\label{section:intro}
% \begin{figure}[t]
%     \centering
%     \includegraphics[width=0.95\columnwidth]{src/流程图.pdf}
%     \caption{Overall process of CaP. \textit{An MWP has a source problem and a target expression. CaP iteratively combine operands and predict operators to generate the target expression layer by layer in a bottom-up manner.}}
%     \label{fig:overall_process}
% \end{figure}

Math Word Problems (MWPs) is a fundamental natural language processing task which requires understanding the natural language description of math problems and inferring the corresponding solution expressions \cite{bobrow1964natural}. %\red{\cite{} Cite the MWP paper here}. 
%The MWP solvers are fed in with a mathematical problem description including several short narratives and a query about the problem, and are expected to generate a corresponding solution expression.
Table \ref{tab:exm_mwp} shows an example: given a mathematical problem description consisting of several short narratives and a query about the problem, the task is predicting the target solution expression, which is essentially equivalent to a tree constituting of math operators and quantities. 

\begin{table}[t]
    \centering
    \begin{tabular}{|l|m{5cm}|}
    \hline
      Source Problem & In one year, there were \textcolor{blue}{76} million voters, whereas in the next year there were \textcolor{blue}{99} million voters. Find the percentage change in the number of voters? \\
      \hline
      Target Expression  & $x=(\textcolor{blue}{99}-\textcolor{blue}{76})\div\textcolor{blue}{76}\div\textcolor{red}{100}$\\
      \hline
    %   Expression Tree & \multicolumn{1}{m{5cm}|}{\centering \includegraphics[width=0.4\columnwidth]{src/exm_tree.pdf}}  \\
    Expression Tree & \multicolumn{1}{m{5cm}|}{\centering \includegraphics[width=4cm]{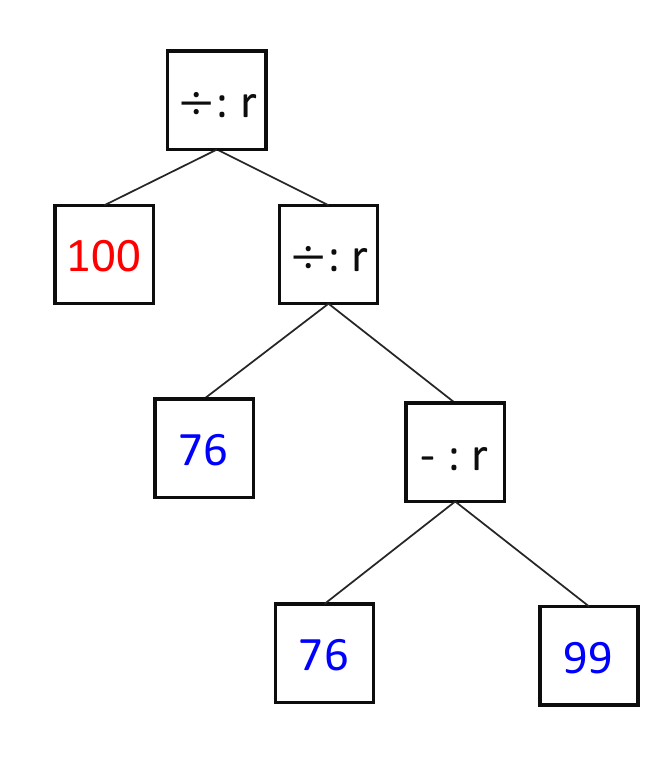}}  \\
    %   \hline
    %   \multicolumn{2}{|c|}{} \\
      \hline
    \end{tabular}
    \caption{An example of MWPs. To normalize, we reorder the operands by the appearance order. Symbols $f$ and $r$ are used to represent the actual calculation order with $f$ indicates the normal order and $r$ is the reverse.}
    \label{tab:exm_mwp}
\end{table}

Previous works mainly formalize MWPs as a generation task and adopt Sequence-to-Sequence (Seq2Seq) \cite{wang2017deep,wang2018translating} or Sequence-to-Tree (Seq2Tree) \cite{xie2019goal,zhang2020graph,wu2020knowledge,wu2021math} models to generate the sequential or tree structure of the solution expression. 
Essentially, solving MWPs is to determine which operands to combine and the operators for the combinations.
However, both kinds of Seq2Seq and Seq2Tree based models directly generate a complete expression from the global representation of the input math problem without 
% However, they all do not 
explicitly learn the combination of operands. 
As a result, these methods will suffer from weaknesses from two aspects: % leading to two major weaknesses: 
i) \textbf{Sensitivity}: 
One of the inherent characteristic of MWPs is sensitivity, which means even minor local variations of the source problem may lead to a totally different solution.
% Local variations in MWPs may lead to a totally different expression.
However, previous methods generating the expression structure in a top-down manner (Seq2Tree) or generating the expression sequentially (Seq2Seq) with a global representation of the input math problem is hard to notice the subtle local variations of the source problem. 
Therefore, they are insensitive to subtle local variations \cite{li2022dataaugmwp} and can easily produce incorrect solutions.
Actually, many of the above-mentioned methods have been proven to learn shallow heuristics purely \cite{patel2021nlp}. 
ii) \textbf{Generalization}: The above-mentioned models have limitations in generating correct expressions for unseen problem types especially when trained with limited data. Considering the target space of MWPs is infinite since mathematical expressions are innumerable, it's impossible to learn all forms of expressions.
% 3) \textbf{Diversity}: In MWP task, one problem can usually be mapped to several diverse but equivalent expressions under mathematical laws. Previous works have tried to address this problem by knowledge distillation \cite{zhangteacher} and equation normalization \cite{wang2018translating}, while our method could alleviate this issue in nature.

To overcome the weaknesses mentioned above, we propose a novel model to hierarchically combine candidate operands and then predict the operators for them in a bottom-up manner. 
% As the name indicates, 
Our proposed model is composed of two modules, namely combination module and prediction module. Combination module is used to make the decision of which operands (referred to as nodes of the expression tree) to combine in pairs. 
% And prediction module predicts the operator and termination status of the generated inner node.
% \red{includes an operator classifier, an inner node encoder and a termination classifier (don't put details in introduction, just talk about the objective of the module)}. 
Given a pair of nodes combined by the combination module, the operator classifier of prediction module will predict the operator for the two operands. 
Besides, a termination probability is calculated to determine whether the generation should be stopped.
% \red{The inner node encoder further encodes the new combination node with the features of two operands and its predicted operator value. Finally, a termination score is predicted with the termination classifier to indicate whether or not the generation should be terminated. (Also too detailed. Such kind of description shall put in the method part.)}
The initial candidate operands nodes of are numbers in the problem and constants with specific meanings (such as constant 100 in Table \ref{tab:exm_mwp} which means percentage). And as the layer increases, the new combination nodes are included in the candidate set iteratively.

% 分别讲什么设计解决了上面的哪一个weakness
Our method outperforms previous methods in the above-mentioned two aspects. 
For \textbf{Sensitivity}, in our method, the decisions of combining operands and the operator, and termination status predictions are all based on the local context features. 
Therefore, our model is more sensitive to local variations.
% we take a bottom-up manner to  combining from the numbers in the problem. And the representation of these numbers are all local context features which are able to capture the local variations. 
% And the whole process of CaP is based on the local features of the problem, so that our model is ensured to be sensitive to local variations. %The experimental results also verify this statement.
For \textbf{Generalization},
different from previous models that learn to directly generate a complete equation (e.g., $x=(a+b)\times(c+d)$), our model learns the combinations (e.g., $y=a+b$, $z=c+d$, and $x=y+z$) separately.
By this means, we make our model learn the most basic compositions instead of the infinite equation sequence. 
Hence, our model has better compositional generalization capability than previous neural models \cite{liu2020compositional}.
% .... Our model could perform more robustly on unseen data.
% 3): equivalent expressions in different forms

Aside from the advantages mentioned above, the method we proposed also has following strengths. First, our method makes full use of the supervision of sub-expressions for each data sample, since all combinations could be restored to a corresponding sub-expression of the target one. 
% And in one iteration, CaP independently learns each combination. 
% since the combinations in one iteration are independent with each other and are calculated loss separately. 
% 太啰嗦了
Second, our model is robust to diverse equivalent expressions. Previous works maximize the joint log-likelihood of the whole expression sequence. Thus, expressions with different forms will be punished even if they are mathematically equivalent. 
Our model is trained to maximize the probability of the target combinations while combinations not appearing in the target expression will be partially punished. By this means, some unlabeled meaningful combinations can also get a high probability. 

% Since combinations in one iteration are independent with each other, making use of the supervision 
% make full use of data
% easier to deal with diverse expressions

% Furthermore, previous works maximize the joint log-likelihood of the whole expression sequence. Thus, expressions with different forms will be punished even if they may be equivalent in mathematical.
% For example, the equation $x_1=a+b+c$ is equivalent with $x_2=a+c+b$ but are treated differently in seq2seq and seq2tree based models. However, our model is trained to maximize the probability of the target combinations while the combinations not occurred in the target expression will not be punished (we adopt hinge-loss to alleviate this issue and more details will be clarified in Sec. \ref{section:method}). 

In summary, our contributions in this paper are: 
\begin{itemize}
    \item We propose a novel model which is more sensitive to the minor local variations. The experimental results show that our method performs better on the challenging data that are similar in text yet with different expressions.
    \item The method we proposed has a better compositional generalization capability and can address unseen problems better.
    \item The experimental results show that our model achieves satisfactory performance on existing benchmark datasets without introducing additional common sense knowledge. 
    % \item We point out that the model frameworks of previous works lack the combination learning which leads to poor sensitivity to subtle local variations and inferior generalization capability. 
    % \item We proposed a novel model CaP which takes substantially different way as existing mainstream MWP neural solvers. 
    % Our model explicitly learns the combination which makes it more sensitive to local variations and has better generalization capability.
    % \item Despite without introducing additional knowledge, CaP achieves competitive and even better performances compared with models augmented with additional knowledge. 
    
    % \item We argue that previous works based on global representation are not local-sensitive. However, our methods based on local features are more sensitive to local variations and can solve MWP more robustly. 
    % \item We propose a brand new model for MWP to separately learn the basic compositions of MWP instead of directly learning the complete equation sequence. Hence CaP has better compositional generalization capability and can perform better on unseen data than previous works.
    % re-check!
    % \item CaP could alleviate the equivalent expressions naturally. ?
    % \item Experimental results on two MWP benchmarks have shown the superiority of CaP over seq2seq and seq2tree based models.
\end{itemize}
\section{Related Works}\label{section:related_works}
The mainstream neural models for MWP contains three kinds of frameworks.

\textbf{Seq2Seq based models}: \cite{wang2017deep} first proposed to adopt recurrent neural network as the encoder and decoder to generate target equations. 
\cite{wang2018translating} further adopted equation normalization to address diverse equivalent expressions. 
\cite{chiang2018semantically} introduced a data structure (i.e. stack) to align the source and target based on Seq2Seq model.
\cite{li2019modeling} brought in attention mechanism and \cite{wang2019template} adopted a two-stage way to generate the sequential expressions.

\textbf{Seq2Tree based models}: \cite{xie2019goal} first proposed to adopt a tree-based LSTM to generate an expression tree instead of a sequence. 
\cite{zhang2020graph,li2020graph} introduced graph structures to better capture features of source problems. 
\cite{zhang2020teacher} utilized knowledge distillation and multiple decoders to generate diverse expressions. 
\cite{wu2020knowledge} introduced common sense knowledge into Seq2Tree based generator. 
And \cite{wu2021math} explicitly encoded the numeric value of numbers into the Seq2Tree model.

\textbf{Pre-trained models}: The experiments in \cite{shen2021generate} have shown that pre-trained language BART \cite{lewis2019bart} could beat many of specifically designed models on Math23k \cite{wang2017deep}. 
Similar results could be found in \cite{lan2021mwptoolkit}. 

\cite{cao2021bottom} proposed a Seq2DAG approach to solve multivariate problems which is similar with our model structure. Different from \cite{cao2021bottom} which enumerated all possible sub-expressions and terminated immediately as the number of generated expressions meets the needs, we design our prediction module more elegantly (refer to Sec.\ref{section:method} for more details). 

The above-mentioned works all ignored the local-sensitivity and compositional generalization ability. Despite \cite{cao2021bottom} also took a bottom-up manner, they still adopted global representation of the problem for inference which makes the model easily overfit the problem types in the training set.
\section{Methodology}\label{section:method}
\subsection{Preliminary}\label{sec:preliminary}
The source problem is a sequence of words with a background description and a query about the desired variable.
% We denote numbers in a MWP problem $Q$ as $N=\{n_0,\cdots,n_l\}$. 
As what have done in previous works, we replace the numbers (denoted as $N$) appearing in the problem $Q$ with a placeholder. % $n_i$ as $i$ indicates the order of appearance in practice. 
% Since the specific numeric value of numbers (denoted as $N$) in the problem $Q$ are , we replace the numbers occurred in the problem with a placeholder $n_i$ as $i$ indicates the order of appearance in practice. 
Besides explicit numbers in $Q$, there are constants with specific meanings also useful for inference, such as $\pi$, $7$ (a week). We denote such constants as $C$.

The target expression can be formalized as a binary tree, of which the leaf nodes are composed of $N\cup C$ and the inner nodes are operators (denoted as $O$). Each node (denoted as $n$) except the leaf nodes in the expression tree can be viewed as a combination of two operands nodes and we refer to it as a combination node or an inner node.
% and thus we denote the composition of the expression tree as node. And if two nodes have the same parent node, we call this parent node a combination node of these two children. 
Each layer of nodes in the expression tree are denoted as $L_i$. %The value of nodes could be numbers $N$, constants $C$, and operators (referred to as $O=\{+,\times,-:f,-:r,\div:f,\div:r\}$). 

Notably, operands calculated with operator $-$ and $\div$ are not commutative (e.g., expression $n_0 - n_1$ is not equivalent with $n_1 - n_0$ in most cases). We adopt $f$ and $r$ to indicate the calculation order of the operands, in which $f$ indicates the normal order while $r$ refers to the reverse. And thus the set of operator should be $O=\{+,\times,-:f,-:r,\div:f,\div:r\}$.
To normalize, when combining operands nodes, we always make the node with smaller index the left operand and vice versa for the right operand.
% in each pair of nodes, we make the node with smaller index the left operand and vice versa for the right operand. 
% In this paper, the nodes are combined iteratively. So the combination nodes must be inner nodes, which means the value of the combination nodes must be $O$.

\begin{figure*}[t]
    \centering
    \includegraphics[width=17.5cm]{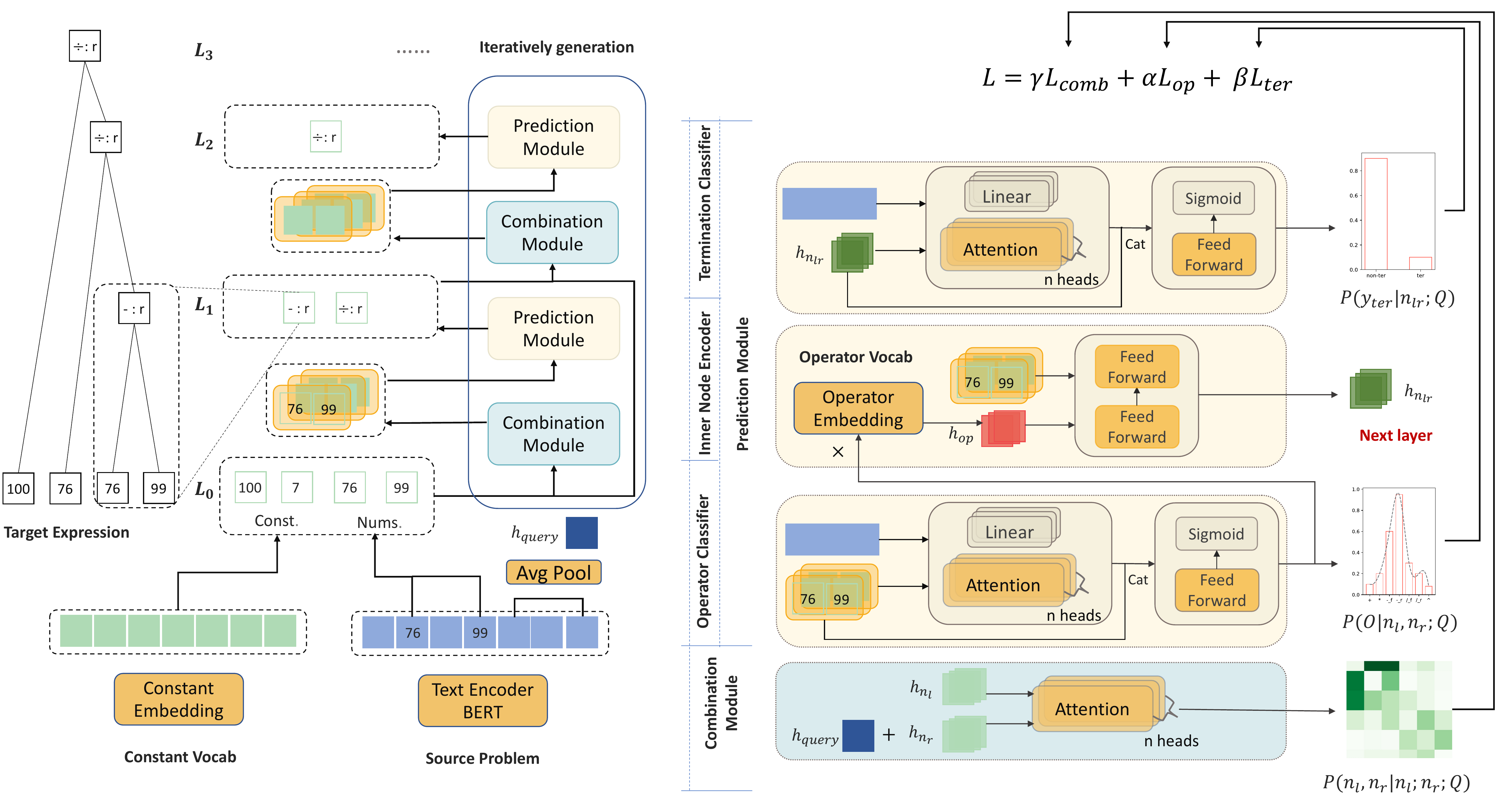}
    \caption{Overview of our proposed method.
    %\textit{The source problems and constant vocabulary are encoded with text encoder and constant embedding module. CaP is composed of combination module and prediction module. Combination module is used to combine candidate operands. Prediction module including operator classifier, inner node encoder and termination classifier is used to predict an operator, encode the new inner node and estimate whether to terminate generation.}
    }
    \label{fig:module_details}
\end{figure*}

\subsection{Overview}\label{sec:overview}
Our method iteratively combines candidate operands nodes from bottom ($L_0$) to up ($L_t$) with two modules, namely combination module and prediction module. 
Combination module is used to determine which nodes to combine in pairs. And prediction module including operator classifier, inner node encoder and termination classifier will further predict the operator and termination status, and also encode the new combination nodes.

% here introduce text encoder, query representation and constant embedding
In the beginning, we adopt a pre-trained language model BERT \cite{devlin2018bert} to encode the problem sequence $Q$. And the hidden states of the number tokens in the last layer of Transformer are used to represent the numbers nodes. 
Respectively, the constants nodes are embedded with a random initialized embedding layer. 
As shown in Fig.\ref{fig:module_details}, numbers $N$ and constants $C$ make up the initial candidate set of operands in $L_0$. 

Besides, as mentioned above, the MWP is composed of a short background description and a query.
Considering that query plays an important role in inference \cite{patel2021nlp}, we utilize average pooling on the tokens in the query to get the representation of the query sub-sentence (denoted as $h_{query}$). 

With the problem encoded and the leaf nodes initialized, the model starts to iteratively generate inner nodes. 
And from $L_0$ to $L_1$, the following steps are took:
1) Combination module combines nodes in $L_0$ in pairs and each with a combination probability. Combination pairs are sent to the prediction module and make up the next layer $L_1$. 
2) The operator classifier in prediction module is used to estimate the probability  $P(O|n_l,n_r;Q)$ given the pair of operands $n_l$ and $n_r$. 
3) And then inner node encoder in prediction module encodes the new inner node $n_{lr}$ with the pair of operands and the operator predicted by operator classifier. 
4) The termination classifier further predicts the termination status which is probability $P(y_{ter}|n_{lr};Q)$ conditioned on the new inner node. 

And then from $L_1$ to $L_2$, step 1 to 4 are repeated. The only difference is that the candidate set for combination has changed from $L_0$ to $L_0 \cup L_1$.

\subsection{Combination Module}\label{sec:comb}
As shown in Fig.\ref{fig:module_details}, the combination module takes two candidate sets as input and combine all possible pairs from these two sets. %The left operands must come from the left candidate set, and the same as the right one. 
The initial candidate sets of left and right operands in $L_0$ are both $N\cup C$. 
As the layer increases, new generated inner nodes are iteratively added to the candidate sets, leading to the exponentially extension of the candidate sets. 
To constrain the scope of nodes in each layer, we adopt beam search mechanism, keeping only top k nodes each layer for layers $L_1$ to $L_t$. 
Besides, when generating new nodes from $L_k$ to $L_{k+1}$, 
%for the next layer (denoted as $L_{k+1}$), 
there must be at least one operand node belongs to layer $L_k$. If both $n_l,n_r\in \cup_{i=0}^{k-1}L_i$, then there could exist an inner node $n_{lr}\in L_k$ which has already combined $n_l$ and $n_r$. %before $L_{k}$ leading to duplication. %$\exists n_{lr}\in L_i, i\leq k-1$ leading to duplication. 
To avoid this issue, from layer $L_k$ to $L_{k+1}$, the left operands are selected from $\cup_{i=0}^{k}L_i$, while the right operands are from $L_k$ only to avoid duplicated combinations.

After combining all possible operands in pairs, we adopt multi-head attention mechanism to predict the combination probability $P(n_l,n_r|n_l;n_r;Q)$ for each pair given the features of the operands and the problem. The probability is calculated with: 
\begin{gather}
    P(n_l,n_r|n_l;n_r;Q)= \notag\\ 
    \sigma (Concat(W_{k_i}^T h_{n_l})^T Concat(W_{q_i}^T (h_{n_r} + h_{query})))
\end{gather}
% \begin{gather}
%     P(n_l,n_r|n_l;n_r;Q)= \notag\\ 
%     \sigma (\mathop{||}\limits_{i=1,\cdots,h}(W_{k_i}^T h_{n_l})^T \odot \mathop{||}\limits_{i=1,\cdots,h}(W_{q_i}^T (h_{n_r} + h_{query})))
% \end{gather}
The $W_{k_i}$ and $W_{q_i}$ are parameters of the i-th head. And we use sigmoid function ($\sigma(\cdot)$) instead of softmax because the combination of pairs is independent with each other.

% loss calculation
For the combination module, there are labeled data and unlabeled data. Combinations appearing in the target equation could be viewed as \textit{positive}, which should be estimated with high combination probability.
However, the combinations not appearing in the target equation are not all \textit{negative}. For example, in equation $x=(a+b)\times c$, nodes pair $(a,c)$ doesn't exist in the target equation, but it is still a meaningful combination. 
So when we calculate loss for unlabeled data, we apply hinge loss to prevent the model from being over-confident on the inaccurate data. The loss function for combination module is:
\begin{small}
\begin{gather}\label{eq:comb_loss}
    L_{comb} = -\sum_{l,r}y_{c_{l,r}}log P(n_l,n_r|n_l;n_r;Q) \notag \\
    + max(\theta, -\sum_{l,r}(1-y_{c_{l,r}})log(1-P(n_l,n_r|n_l;n_r;Q)))
\end{gather}
\end{small}

$y_{c_{l,r}}$ here refers to whether combination of nodes $n_l$ and $n_r$ appearing in the target equation, and $\theta$ is a hyper-parameter.

\subsection{Prediction Module}
Prediction module is composed of three parts: an operator classifier to predict the operator for the input combinations; an inner node encoder to encode the generated combination node; and a termination classifier to determine whether to terminate the generation.
\subsubsection{Operator Classifier}\label{sec:operator}
Given a pair of operands, the operator classifier outputs an operator probability $P(O|n_l,n_r;Q)$,
\begin{align}
    P(O|n_l,n_r;Q) &= \sigma (FF([c_O;h_{n_l};h_{n_r}])) \\
    c_O &= Attention(Q,[h_{n_l};h_{n_r}])
\end{align}
$FF(\cdot)$ represents the feed forward network and $[]$ refers to concatenation function. $c_O$ is the context representation obtained by adopting attention mechanism on nodes and problem sequence. The hidden vector of nodes are viewed as query while the problem sequence is the key and value. Details of the attention mechanism could refer to \cite{vaswani2017attention}. Besides, we choose sigmoid as the activate function instead of softmax here because one pair of operands may have multiple operators. 

The objective of operator classifier is to minimize the negative log-likelihood: 
\begin{gather}
    L_{op}=\sum_{o\in O; n_l,n_r}-y_{o_{l,r}}log P(y=o|n_l,n_r;Q) \notag \\
    - (1-y_{o_{l,r}})log (1-P(y=o|n_l,n_r;Q))
\end{gather}
$y_{o_{l,r}}$ indicates whether node $n_l$ and $n_r$ has a parent node whose value is operator $o$ in the ground truth expression.

\subsubsection{Inner Node Encoder}\label{sec:inner}
As shown in Fig.\ref{fig:module_details}, after predicting proper operators for the operands pairs, inner node encoder encodes the operands and the predicted operator to obtain the representation of the combination nodes. 
The obtained combination nodes making up the next layer of generated expression trees are then added to the candidate sets. 

The representation of combination nodes are calculated as follows:
\begin{align}
    h_{n_{lr}} &= FF([h_{n_l};h_{n_r};h_{op}]) \\
    h_{op} &= \sum_{o\in O}P(y=o|n_l,n_r;Q)e_o
\end{align}
$FF(\cdot)$ refers to feed forward network and $e_o$ is the trainable embedding vector of operator $o$ which is initialized randomly.

\subsubsection{Termination Classifier}\label{sec:termination}
Since the expression cannot be generated endlessly, a termination classifier is utilized to predict a termination probability $P(y_{ter}|n_{lr};Q)$ for each combination node $n_{lr}$. %given the representation of the combination node obtained from inner node encoder and the features of the problem sequence. 
Similar to prediction classifier, we adopt attention mechanism to get a context representation $c_t$. 
The probability of termination is formulized as: 
\begin{align}
    P(y_{ter}|n_{lr};Q) &= Softmax (FF([c_t;h_{n_{lr}}])) \\
    c_t &= Attention(Q,h_{n_{lr}})
\end{align}
$y_{ter}=1$ indicates the expression should be terminated,
% The label of termination is 0 for non-termination and 1 for termination. 
Only the root node of the ground truth expression is labeled as terminated, and the other inner nodes are all non-terminated nodes. 
The loss function of termination classifier is:
\begin{gather}
    L_{ter}=-y_{ter}log P(y_{ter}|n_{lr};Q) \notag \\
    - (1-y_{ter})log(1-P(y_{ter}|n_{lr};Q))
\end{gather}

\subsection{Training ang Inference Procedure}
For simplification, only combinations appeared in the target expression trees are sent to the prediction module to make up the next layer of the tree during training phase.
The entire objective of our method is to minimize the loss of three parts: 
\begin{gather}
    L=\gamma L_{comb} + \alpha L_{op} + \beta L_{ter}
\end{gather}
$\alpha, \beta, \gamma$ are all hyper-parameters.

When performing inference, in every iteration, combinations with top k highest combination probability are selected as the nodes of next layer. 
We also set $L_{max}$ (i.e. the max layer to generate) to prevent endless generation.
Besides, choosing root nodes only based on termination probability lacks the information of combination and operator confidence. So for each node, we calculate a joint probability of its descendants about combination and operator probability.
The joint probability is defined as below:
\begin{small}
\begin{equation*}
    P_{\mathcal{J}}(n_i)=\left\{
    \begin{aligned}
        &0 & n_i\in L_0 \\
        &P_{op}(n_i)*P_{comb}(n_i) & n_i \in L_1 \\
        % \frac{(P_{\mathcal{J}}(l(n_i)) + P_{\mathcal{J}}(r(n_i)) )*P_{op}(n_i)*P_{comb}(n_i)}{Count(l(n_i),r(n_i)\notin L_0)} & otherwise
        &Avg(P_{\mathcal{J}}(\mathcal{C}(n_i)))*P_{op}(n_i)*P_{comb}(n_i) & others
    \end{aligned}
    \right.
\end{equation*}
\end{small}
$P_{comb}$ and $P_{op}$ are introduced in Sec.\ref{sec:comb} and \ref{sec:operator} respectively. $\mathcal{C}(n_i)$ refers to the direct left and right child of node $n_i$. Function $Avg(\cdot)$ is used to calculate the average joint probability of the left and right child. Notably, the child node with a joint probability of 0 is not considered when calculating the average probability. 
And the final probability of $n_i$ being the root node is obtained by multiplying its termination probability with the joint probability:
\begin{gather}
P_{root}(n_i)=P_{\mathcal{J}}(n_i)*P_{ter}(n_i)
\end{gather}
At last, the node with the highest $P_{root}$ is chosen as the root node to restore a predicted expression tree. 
\section{Experiments}\label{section:experiments}
\subsection{Experimental Setup}\label{sec:experimental_setup}
% \subsubsection{Datasets}
\textbf{Datasets} In this paper, we conduct our experiments on two benchmark datasets: Math23k \cite{wang2017deep} and SVAMP \cite{patel2021nlp}, and two transformed datasets: Math23k+LGA \cite{li2022dataaugmwp} and re-split SVAMP. 

% three datasets: , Math23k+LGA (cite semantic-based data augmentation) and SVAMP \cite{patel2021nlp}. 
Benchmark datasets: 1) Math23k is one of the most popular MWP datasets with 23,161 data samples in Chinese. 
% Besides, to prove that our method is more sensitive to local variations than previous works, we also conduct experiments on Math23k+LGA. 
2) SVAMP is a more challenging dataset which is built by applying different kinds of subtle variations on problems from ASDiv-A \cite{miao-etal-2020-diverse}. 
% SVAMP build a challenging dataset by applying different kinds of subtle variations on problems from ASDiv-A \cite{miao-etal-2020-diverse}. 
The training set of it is a combination of MAWPS \cite{koncel2016mawps} and ASDiv-A containing only simple problems with 3,138 data samples. 
And the testing set is composed of 1k varied unseen problems which is very challenging for existing neural models.
We also perform 5-fold cross-validation over SVAMP as in \cite{patel2021nlp}. The dataset is divided into the training set containing data from MAWPS, AsDiv-A and some varied problems, and the testing set containing the rest of varied problems.
% Our method is evaluated with 5-fold cross-validation as in \cite{patel2021nlp}, the size of the training set is approximately 3.9k and the testing set is about 200 samples.

Transformed datasets: 1) The training set of Math23k + LGA is Math23k training set augmented by LGA. The augmentation brings data with tiny local variations into the new training set. 
The testing set of Math23k+LGA is the same as Math23k.
% The original training set of Math23k is augmented with LGA (cite semantic-based data augmentation) which introduces data with tiny local variations while the testing set remains the same as Math23k.
% And since experiments have shown that evaluation results on Math23k is not able to prove the inference ability of neural models \cite{patel2021nlp}(cite semantic-based data augmentation), we also conduct experiments on SVAMP, a more challenging dataset.
2) We re-split SVAMP by equation templates\footnote{Equation templates are obtained by converting the equations into prefix form and masking out all numbers with a meta symbol.\cite{patel2021nlp}}. 
The training set contains data samples with 10 equation templates. 
And equation templates of the testing data are all unseen. 
% Equations with one operator and several other random picked templates are splited into the new training set. Data in the testing set contains only unseen equation templates.

The evaluation metric on all datasets is answer accuracy, which measures whether or not the calculated answer of the predicted equation is the same as the ground truth answer.

% \subsubsection{Baselines}
\textbf{Baselines} We compare our models with various typical Seq2Seq-based, Seq2Tree-based and pre-trained methods including: 
1) DNS \cite{wang2017deep}: a vanilla Seq2Seq model, Transformer \cite{vaswani2017attention}.
2) GTS \cite{xie2019goal}: a GRU encoder and a tree-structured LSTM decoder. 
Graph2Tree \cite{zhang2020graph}: several pre-defined graphs with GCN to encode and a tree-structured LSTM decoder.
3) BERT2Seq \cite{lan2021mwptoolkit}: BERT as encoder and a Transformer decoder. 
4) Seq2DAG \cite{cao2021bottom}: a bottom-up generator with DAG-LSTMs as the decoder.
5) Generate\&Rank \cite{shen2021generate}: a SOTA framework of first generating top k candidates with a generator BART \cite{lewis2019bart} and then ranking them to get the final result. 
% BART \cite{lewis2019bart}: a pre-trained Seq2Seq model which is used as an expression generator in Generate\&Rank .

% Considering we adopt a pre-trained language model BERT to encode the source problem in CaP, to eliminate the performance gains brought in by BERT, we also compare with Bert+model on SVAMP dataset and Bert2Seq on Math23k dataset. \textcolor{red}{RECHECK!}

% \subsubsection{Implementation details}
% The algorithm is implemented with PyTorch\footnote{https://pytorch.org/} and \red{the pre-trained language model is Transformers\footnote{https://huggingface.co/}}.
\textbf{Implementation Details} The dimensions of constant and operator embedding are set to 256. 
Hidden dimensions of all feed forward network and attention parameters are set to 512. 
2-layer feed forward network is used in operator and termination classifier with ReLU as activation function. 
% For Math23k dataset, we use 3-layer feed forward network for operator and termination classifier. And for SVAMP, we adopt 2-layer feed forward network to avoid overfitting. 
$L_{max}$ and the size of beam search $k$ are set to the max number of layers and nodes per layer counted from the training set respectively.
We adopt Adam optimizer \cite{kingma2014adam} with the initial learning rate set to 4e-5 and 8e-5 for SVAMP and Math23k respectively, and the learning rate is halved every 20 epochs. 
The hyper-parameters of loss weights are set as $\alpha=1.5, \beta=1.0, \gamma=0.05$ to balance the loss from different sub-modules.
And $\theta$ in combination loss is set to 1.5 for Math23k and 1.3 for SVAMP since the beam size $k$ in SVAMP is smaller and thus fewer negative samples will be generated.
The batch size for the two datasets are 64 and 256 considering SVAMP is too small for large batch size. 
We train our model on both datasets for 200 epochs. 
All experiments are carried on 4 NVIDIA RTX 3090. 

\begin{table}[t]
    \centering
    \begin{tabular}{l|c|c|c|c}
        \toprule[1.5pt]
        \multirow{3}{2.3cm}{Model}  & \multicolumn{4}{c}{Dataset} \\ \cline{2-5}
        & \multicolumn{2}{c|}{SVAMP} & \multicolumn{2}{c}{Math23k} \\ \cline{2-5}
         & Full & 5-fold & Full & + LGA  \\
         \midrule[0.8pt]
         \multicolumn{5}{l}{Seq2Seq based models} \\
         \midrule[0.8pt]
         DNS & 24.2 & 29.5 & 66.1 & 71.2  \\
         Transformer & 20.7 & 22.7 & 61.5 & 64.0 \\
         \midrule[0.8pt]
         \multicolumn{4}{l}{Seq2Tree based models} \\
         \midrule[0.8pt]
         GTS & 30.8 & 38.8 & 75.6 & 76.1 \\
        %  GTS+bert & 32.9 & 49.2 & 82.0 &  \\
         Graph2Tree & \textbf{36.5} & 47.8 & 77.4 & 77.0 \\
        %  Graph2Tree+bert & \textbf{40.3} & 54.7 & 76.5 &\\
         \midrule[0.8pt]
         \multicolumn{4}{l}{pre-trained models} \\
         \midrule[0.8pt]
         BERT2Seq & 24.8 & 51.5 & 76.6 &  76.0\\
         \midrule[0.8pt]
         Seq2DAG \footnotemark[3] & - & - & 77.1 & -\\
         \midrule[0.8pt]
         BART & -&- & 80.8 &-  \\
         Generate\&Rank\tablefootnote{Since the code of hasn't been released yet, we only report the results in the original paper.} &- & - & \textbf{85.4} & - \\
         \midrule[0.8pt]
         Ours & 35.1 & \textbf{57.4} & 79.9 & 82.1  \\
        \bottomrule[1.5pt]
    \end{tabular}
    \caption{Answer accuracy (\%) of baseline models and our method on the benchmark datasets and a transformed dataset Math23k+LGA. 
    The best performances are marked with bold.}
    \label{tab:main_experiment}
\end{table}

\begin{table}[t]
    \centering
    \begin{tabular}{l|c|c}
       \toprule[1.5pt]
       Model  & Ans acc. (\%) & Predicted unseen templates \\
       \midrule[0.8pt]
       DNS & 10.1 & 3 \\
       GTS  & 12.1 & 0 \\
       Graph2Tree & 11.9 & 6 \\
       \midrule[0.8pt]
       Ours & \textbf{17.1} & \textbf{33} \\
       \bottomrule[1.5pt]
    \end{tabular}
    \caption{Experimental results on the re-split SVAMP. %Ans acc. shows the answer accuracy of the neural models. 
    \textit{Predicted unseen templates} refers to the number of samples that are predicted an unseen equation template in the validation set. 
    The best results are marked as bold.}
    \label{tab:generalization_exp}
\end{table}

\subsection{Results and Analysis}
We conduct experiments with four different settings on the two benchmark datasets as shown in Table \ref{tab:main_experiment}. 
Full set of both SVAMP and Math23k contains only simple problems in the training set. But the testing set of SVAMP-Full contains various challenging data that are similar with each other in text yet different in semantics, while the testing set of Math23k-Full contains only simple problems.
The 5-fold data sets of SVAMP are not simply randomly divided as mentioned in Sec.\ref{sec:experimental_setup}. Both training and testing sets of the 5-fold cross validation contain challenging samples. 
Similarly, the training set of Math23k+LGA contains massive challenging samples with tiny local variations.

% \subsubsection{Overall analysis}
\textbf{Overall Analysis} Notably, the ideas of Generate\&Rank and our model are orthogonal. Generate\&Rank proposes a framework of first generating top k candidates with a generator (BART in their original paper) and then ranking the candidates. 
Our model naturally generate various candidate expression trees for each problem and can be used as a generator in the Generate\&Rank framework.
% is a generator which could be further applied with this framework. 
Despite our model does not beat Generate\&Rank on Math23k dataset, Our method outperforms the generator (BART) they used with 82.1\% answer accuracy (the testing set of Math23+LGA is the same with Math23k-Full).

By comparing the results of Bert2Seq and our method shown in Table \ref{tab:main_experiment}, our model beats the result of BERT2Seq on all experimental settings, which shows that the performance gains of our model not only comes from the pre-trained language model but also from the specifically designed model architecture. 

The results on SVAMP-Full indicate that when trained without those challenging samples, almost all neural models lack the capacity of dealing with challenging problems with tiny local variations as the performances of all neural models on SVAMP-Full are quite unsatisfactory. 
Notably, additional information like the pairwise comparative relationship between numbers are given in Graph2Tree, which is helpful to decide whether to generate $a-b$ or $b-a$. And the testing set of SVAMP is filled with this kind of problems, so that Graph2Tree performs better on SVAMP than all other models. 
Even without introducing additional knowledge, our model is able to achieve a competitive performance as Graph2Tree, which to some extent verify the local sensitivity of our method.
% Among the methods not introducing additional knowledge, CaP achieves a competitive result as GTS which achieves the highest accuracy on SVAMP-Full.

% Besides, on the Math23k-Full dataset, CaP is able to beat Graph2Tree with 2.5\% increment. 

% Besides, the experimental results show that naive pre-trained language models could learn simple problems better than specifically designed Seq2Tree models as the results on Math23k-Full have shown. 
% And with the model architecture we proposed, 
% And the architecture we proposed in this paper also benefits the learning of simple problems. 

% \subsubsection{Local sensitivity to the source problem}
\textbf{Local Sensitivity to Source Problems} As described above, the experimental results on the 5-fold data sets of SVAMP directly reflect the local-sensitivity of neural models since both training and testing sets in this setting contains challenging problems.
% As shown in Tab.\ref{tab:main_experiment}, the performances on SVAMP directly reflects the local-sensitivity of neural models since problems in SVAMP are similar with each other in text yet different in semantics. 
Our method outperforms previous works including Seq2Seq, Seq2Tree and pre-trained models based methods on 5-fold cross validation over SVAMP. 

Besides, the performances of GTS, Graph2Tree and BERT2SEQ descend when trained with the augmented challenging data (the accuracy on Math23k+LGA is lower than that on Math23k-Full). As \cite{li2022dataaugmwp} explained that the descendants are cased by previous neural models not good at dealing with subtle variations. 
However, our method gains significant improvements from the augmented data due to the local-sensitive nature of our method.
% because our model is more sensitive to local variations of the source problems.

% \subsubsection{Generalization ability on unseen data}
\textbf{Generalization Ability on Unseen Data} Table \ref{tab:generalization_exp} shows the experimental results on the re-split SVAMP. The performances of all models are inferior on this dataset due to the difficulty. 
But among them, our method achieves the best performance. 
Besides the accuracy, the capacity of our method predicting unseen equation templates significantly outperforms previous works including both Seq2Seq and Seq2Tree models. 

\begin{figure}
    \centering
    \includegraphics[width=0.95\columnwidth]{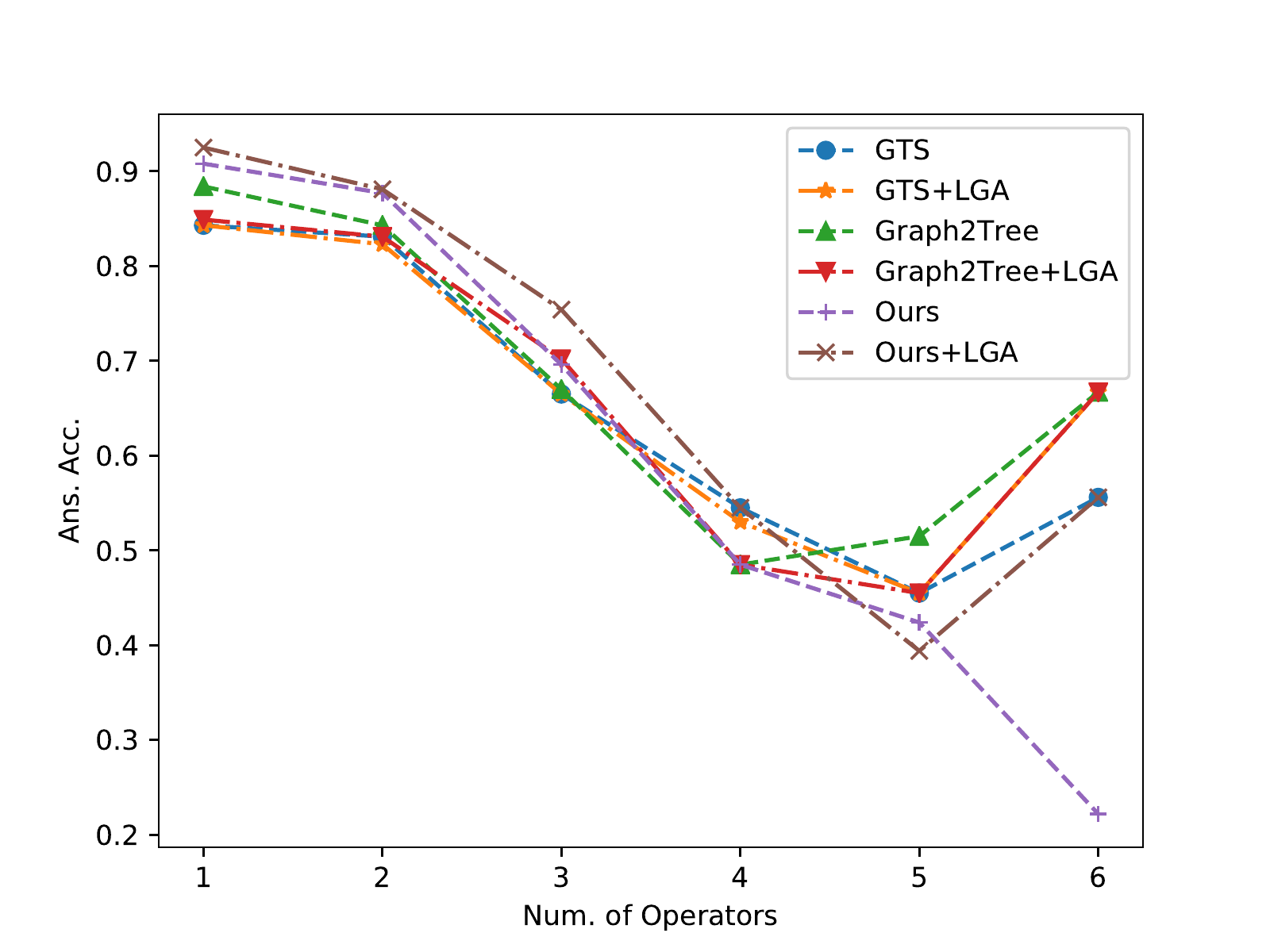}
    \caption{Performances verses equation complexity.}
    \label{fig:acc_complexity}
\end{figure}

% \subsubsection{Performances verses complexity of equations}\label{sec:acc_vs_complexity}
\textbf{Performances verses Complexity of Equations} We count the accuracy for problems with different number of operators as shown in Fig.\ref{fig:acc_complexity}. 
Our method performs better than other models on problems with fewer operators. 
We believe this is because by separately learning the combinations, both problems with long equations and short equations benefit the learning of simple problems. So the learnable data for our model of fewer combinations is much more than previous models which directly learn the complete equations.
However, as combination nodes in higher layers gradually forget the semantics of nodes in lower layers, the performances of our model on longer equations are not quite satisfactory.
% However, for longer equations, 
% the semantics of leaf layer is gradually diluted and the errors are accumulated leading to a poor performance.

% \subsection{Case Study}
% \begin{table}[t]
%     \centering
%     \begin{tabular}{c|c}
%          &  \\
%          & 
%     \end{tabular}
%     \caption{Caption}
%     \label{tab:my_label}
% \end{table}
\section{Conclusion}\label{section:conclusion}
% The lack of shortcut mechanism makes it harder for higher layers to notice the semantics in lower layers. And this limitation makes the performance of CaP inferior on problems with longer equations. 
% Besides, unlike previous methods, we don't introduce additional common sense knowledge which is helpful for inference. 
% 缺少short cuts导致层数较高后会逐渐缺失底层语义信息
% 一定要提自己的方法没有像Graph2Tree一样引入额外的知识，在没有引入知识的模型中取得了和seq2tree-based model差不多的结果，并在一些方面比它的效果更好
%现有的模型都可以再进步，考虑到他们在challenging dataset和unseen problems上表现都还不太好

In this paper, we propose a novel fine-to-coarse model which iteratively combines the candidate operands and predict the operators. Different from previous Seq2Seq or Seq2Tree based models, our model is more local-sensitive and has better generalization capability. 
Experimental results have shown the effectiveness of the model architecture we proposed.

There are still a lot to do in the future works. For example, we will introduce shortcut mechanism to avoid forgotten of combination nodes in higher layers.

\bibliographystyle{named}
\bibliography{citation}

\begin{thebibliography}{}

\bibitem[\protect\citeauthoryear{Bobrow}{1964}]{bobrow1964natural}
Daniel~G Bobrow.
\newblock Natural language input for a computer problem solving system.
\newblock 1964.

\bibitem[\protect\citeauthoryear{Cao \bgroup \em et al.\egroup
  }{2021}]{cao2021bottom}
Yixuan Cao, Feng Hong, Hongwei Li, and Ping Luo.
\newblock A bottom-up dag structure extraction model for math word problems.
\newblock In {\em Proc. of AAAI}, 2021.

\bibitem[\protect\citeauthoryear{Chiang and
  Chen}{2018}]{chiang2018semantically}
Ting-Rui Chiang and Yun-Nung Chen.
\newblock Semantically-aligned equation generation for solving and reasoning
  math word problems.
\newblock {\em arXiv preprint arXiv:1811.00720}, 2018.

\bibitem[\protect\citeauthoryear{Devlin \bgroup \em et al.\egroup
  }{2018}]{devlin2018bert}
Jacob Devlin, Ming-Wei Chang, Kenton Lee, and Kristina Toutanova.
\newblock Bert: Pre-training of deep bidirectional transformers for language
  understanding.
\newblock {\em arXiv preprint arXiv:1810.04805}, 2018.

\bibitem[\protect\citeauthoryear{Kingma and Ba}{2014}]{kingma2014adam}
Diederik~P Kingma and Jimmy Ba.
\newblock Adam: A method for stochastic optimization.
\newblock {\em arXiv preprint arXiv:1412.6980}, 2014.

\bibitem[\protect\citeauthoryear{Koncel-Kedziorski \bgroup \em et al.\egroup
  }{2016}]{koncel2016mawps}
Rik Koncel-Kedziorski, Subhro Roy, Aida Amini, Nate Kushman, and Hannaneh
  Hajishirzi.
\newblock Mawps: A math word problem repository.
\newblock In {\em Proc. of ACL}, 2016.

\bibitem[\protect\citeauthoryear{Lan \bgroup \em et al.\egroup
  }{2021}]{lan2021mwptoolkit}
Yihuai Lan, Lei Wang, Qiyuan Zhang, Yunshi Lan, Bing~Tian Dai, Yan Wang,
  Dongxiang Zhang, and Ee-Peng Lim.
\newblock Mwptoolkit: An open-source framework for deep learning-based math
  word problem solvers.
\newblock {\em arXiv preprint arXiv:2109.00799}, 2021.

\bibitem[\protect\citeauthoryear{Lewis \bgroup \em et al.\egroup
  }{2019}]{lewis2019bart}
Mike Lewis, Yinhan Liu, Naman Goyal, Marjan Ghazvininejad, Abdelrahman Mohamed,
  Omer Levy, Ves Stoyanov, and Luke Zettlemoyer.
\newblock Bart: Denoising sequence-to-sequence pre-training for natural
  language generation, translation, and comprehension.
\newblock {\em arXiv preprint arXiv:1910.13461}, 2019.

\bibitem[\protect\citeauthoryear{Li \bgroup \em et al.\egroup
  }{2019}]{li2019modeling}
Jierui Li, Lei Wang, Jipeng Zhang, Yan Wang, Bing~Tian Dai, and Dongxiang
  Zhang.
\newblock Modeling intra-relation in math word problems with different
  functional multi-head attentions.
\newblock In {\em Proc. of ACL}, 2019.

\bibitem[\protect\citeauthoryear{Li \bgroup \em et al.\egroup
  }{2020}]{li2020graph}
Shucheng Li, Lingfei Wu, Shiwei Feng, Fangli Xu, Fengyuan Xu, and Sheng Zhong.
\newblock Graph-to-tree neural networks for learning structured input-output
  translation with applications to semantic parsing and math word problem.
\newblock {\em arXiv preprint arXiv:2004.13781}, 2020.

\bibitem[\protect\citeauthoryear{Li \bgroup \em et al.\egroup
  }{2022}]{li2022dataaugmwp}
Ailisi Li, Jiaqing Liang, and Yanghua Xiao.
\newblock Semantic-based data augmentation for math word problems.
\newblock {\em arXiv preprint arXiv:2201.02489}, 2022.

\bibitem[\protect\citeauthoryear{Liu \bgroup \em et al.\egroup
  }{2020}]{liu2020compositional}
Qian Liu, Shengnan An, Jian-Guang Lou, Bei Chen, Zeqi Lin, Yan Gao, Bin Zhou,
  Nanning Zheng, and Dongmei Zhang.
\newblock Compositional generalization by learning analytical expressions.
\newblock {\em arXiv preprint arXiv:2006.10627}, 2020.

\bibitem[\protect\citeauthoryear{Miao \bgroup \em et al.\egroup
  }{2020}]{miao-etal-2020-diverse}
Shen-yun Miao, Chao-Chun Liang, and Keh-Yih Su.
\newblock A diverse corpus for evaluating and developing {E}nglish math word
  problem solvers.
\newblock In {\em Proc. of ACL}, 2020.

\bibitem[\protect\citeauthoryear{Patel \bgroup \em et al.\egroup
  }{2021}]{patel2021nlp}
Arkil Patel, Satwik Bhattamishra, and Navin Goyal.
\newblock Are nlp models really able to solve simple math word problems?
\newblock {\em arXiv preprint arXiv:2103.07191}, 2021.

\bibitem[\protect\citeauthoryear{Shen \bgroup \em et al.\egroup
  }{2021}]{shen2021generate}
Jianhao Shen, Yichun Yin, Lin Li, Lifeng Shang, Xin Jiang, Ming Zhang, and Qun
  Liu.
\newblock Generate \& rank: A multi-task framework for math word problems.
\newblock {\em arXiv preprint arXiv:2109.03034}, 2021.

\bibitem[\protect\citeauthoryear{Vaswani \bgroup \em et al.\egroup
  }{2017}]{vaswani2017attention}
Ashish Vaswani, Noam Shazeer, Niki Parmar, Jakob Uszkoreit, Llion Jones,
  Aidan~N Gomez, {\L}ukasz Kaiser, and Illia Polosukhin.
\newblock Attention is all you need.
\newblock In {\em Advances in neural information processing systems}, 2017.

\bibitem[\protect\citeauthoryear{Wang \bgroup \em et al.\egroup
  }{2017}]{wang2017deep}
Yan Wang, Xiaojiang Liu, and Shuming Shi.
\newblock Deep neural solver for math word problems.
\newblock In {\em Proc. of EMNLP}, 2017.

\bibitem[\protect\citeauthoryear{Wang \bgroup \em et al.\egroup
  }{2018}]{wang2018translating}
Lei Wang, Yan Wang, Deng Cai, Dongxiang Zhang, and Xiaojiang Liu.
\newblock Translating a math word problem to an expression tree.
\newblock {\em arXiv preprint arXiv:1811.05632}, 2018.

\bibitem[\protect\citeauthoryear{Wang \bgroup \em et al.\egroup
  }{2019}]{wang2019template}
Lei Wang, Dongxiang Zhang, Jipeng Zhang, Xing Xu, Lianli Gao, Bing~Tian Dai,
  and Heng~Tao Shen.
\newblock Template-based math word problem solvers with recursive neural
  networks.
\newblock In {\em Proc. of AAAI}, 2019.

\bibitem[\protect\citeauthoryear{Wu \bgroup \em et al.\egroup
  }{2020}]{wu2020knowledge}
Qinzhuo Wu, Qi~Zhang, Jinlan Fu, and Xuan-Jing Huang.
\newblock A knowledge-aware sequence-to-tree network for math word problem
  solving.
\newblock In {\em Proc. of EMNLP}, 2020.

\bibitem[\protect\citeauthoryear{Wu \bgroup \em et al.\egroup
  }{2021}]{wu2021math}
Qinzhuo Wu, Qi~Zhang, Zhongyu Wei, and Xuan-Jing Huang.
\newblock Math word problem solving with explicit numerical values.
\newblock In {\em Proc. of ACL}, 2021.

\bibitem[\protect\citeauthoryear{Xie and Sun}{2019}]{xie2019goal}
Zhipeng Xie and Shichao Sun.
\newblock A goal-driven tree-structured neural model for math word problems.
\newblock In {\em Proc. of IJCAI}, 2019.

\bibitem[\protect\citeauthoryear{Zhang \bgroup \em et al.\egroup
  }{2020a}]{zhang2020teacher}
Jipeng Zhang, Ka~Wei LEE, Ee-Peng Lim, Wei Qin, Lei Wang, Jie Shao, Qianru Sun,
  et~al.
\newblock Teacher-student networks with multiple decoders for solving math word
  problem.
\newblock 2020.

\bibitem[\protect\citeauthoryear{Zhang \bgroup \em et al.\egroup
  }{2020b}]{zhang2020graph}
Jipeng Zhang, Lei Wang, Roy Ka-Wei Lee, Yi~Bin, Yan Wang, Jie Shao, and Ee-Peng
  Lim.
\newblock Graph-to-tree learning for solving math word problems.
\newblock In {\em Proc. of ACL}, 2020.

\end{thebibliography}

\end{document}